\documentclass[preprint,12pt]{elsarticle}

\usepackage{graphicx}
\graphicspath{{./Figures/}}
\usepackage{subcaption}
\usepackage{amssymb}
\usepackage{textcomp}
\usepackage{gensymb}
\usepackage{amsmath}
\usepackage{amsthm}
\usepackage{array}
\usepackage{multirow}
\usepackage{wrapfig}
\usepackage{xcolor}
\usepackage{lineno}
\biboptions{sort&compress}

\journal{Extreme Mechanics Letters}

\begin{document}

\begin{frontmatter}

\title{Exploiting the Nonlinear Stiffness of TMP Origami Folding to Enhance Robotic Jumping Performance}
\author{Sahand Sadeghi$^1$ \fnref{label1} \fnref{label2}}
\author{Samuel R. Allison \fnref{label2}} 
\author{Blake Bestill}
\author{and Suyi Li}

\fntext[label1]{Corresponding Author: Sahand Sadeghi \texttt{ssadegh@g.clemson.edu}}
\fntext[label2]{Equal contribution}

\address{Department of Mechanical Engineering\\Clemson University, SC, USA}

%\vspace{-2.0in}

\begin{abstract}
Via numerical simulation and experimental assessment, this study examines the use of origami folding to develop robotic jumping mechanisms with tailored nonlinear stiffness to improve dynamic performance.  Specifically, we use Tachi-Miura Polyhedron (TMP) bellow origami  --- which exhibits a nonlinear ``strain-softening'' force-displacement curve --- as a jumping robotic skeleton with embedded energy storage.  TMP's nonlinear stiffness allows it to store more energy than a linear spring and offers improved jumping height and airtime.  Moreover, the nonlinearity can be tailored by directly changing the underlying TMP crease geometry.  A critical challenge is to minimize the TMP's hysteresis and energy loss during its compression stage right before jumping.  So we used the plastically annealed lamina emergent origami (PALEO) concept to modify the TMP creases.  PALEO increases the folding limit before plastic deformation occurs, thus improving the overall strain energy retention.  Jumping experiments confirmed that a nonlinear TMP mechanism achieved roughly 9\% improvement in air time and a 13\% improvement in jumping height compared to a ``control'' TMP sample with a relatively linear stiffness. This study's results validate the advantages of using origami in robotic jumping mechanisms and demonstrate the benefits of utilizing nonlinear spring elements for improving jumping performance.  Therefore, they could foster a new family of energetically efficient jumping mechanisms with optimized performance in the future.
\end{abstract}

\begin{keyword}
Origami \sep Jumping Mechanism \sep Nonlinear Stiffness 
\end{keyword}

\end{frontmatter}

%\linenumbers

\section{Introduction}\label{Section:1}

The ongoing advances in robotics technology have allowed for exponential growth in their applications in many industries.  Robots have been critical to the recent growth of large-scale manufacturing \cite{shneier}, as well as the energy, food, agriculture \cite{shukla, bogue, chua, reddy}, education \cite{benitti,berghe}, and medical care sectores \cite{jamwal, robinson}. Additionally, the ability to send robots into hazardous environments in place of humans has shown tremendous advantages in military applications \cite{blokhin,kunchev}, and search and rescue operations \cite{wang, reddy2}.  These robots sent into the field often encounter challenging terrains with unpredictable obstacles, which require extensive research into various modes of locomotion for navigation \cite{zhuang}.  To this end, we have witnessed many studies of ground-contact-based locomotion, separated into five main categories: wheeled robots, tracked robots, snake robots, legged robots, and wheel-legged robots \cite{zhuang}. While each category has distinct advantages in specific applications, legged robots are of particular interest in rough terrains due to their ability to gain discrete footholds on various surfaces and traverse steep inclines \cite{zhuang, dubey, cherouvim}. 

Among the various types of legged robots, jumping robots are particularly interesting because locomotion through jumping has several merits compared to walking or crawling.  For example, jumping has higher energy density, efficient obstacle negotiation, and rapid terrain transition \cite{zhang2020biologically, gvirsman2016dynamics, siwanowicz2017three}. Jumping robots are capable of moving through different heights and irregular terrains, and can achieve more convenient freight transportation \cite{zhang2020biologically}, patient care \cite{zhang2020biologically}, disaster relief \cite{wei2014research}, rescue \cite{tsukagoshi2005design}, and interstellar exploration \cite{zhang2020biologically}.  

For high-performance jumping, energy storage is crucial \cite{zhao, sadeghi}, so researchers have investigated various energy storing and releasing mechanisms. Besides the traditional spring \cite{scarfogliero, zhang, zhao2} and pneumatic devices \cite{tsukagoshi}, customized nonlinear mechanisms \cite{dubowsky, yamada} were also examined for their unique benefits to the overall jumping performance. For example, Yamada \textit{et. al.} exploited the snap-through buckling of a closed elastica to store greater amounts of energy for increased jumping height and distance \cite{yamada,yamada2}.  Fiorini \textit{et. al.} designed a 6-bar linkage coupled with gears at the joints with a nonlinear force response curve and investigated its effects on overcoming the pre-mature take-off of the jumper \cite{fiorini}.  

\begin{figure}[h!]
    \centering
    \includegraphics[scale=1]{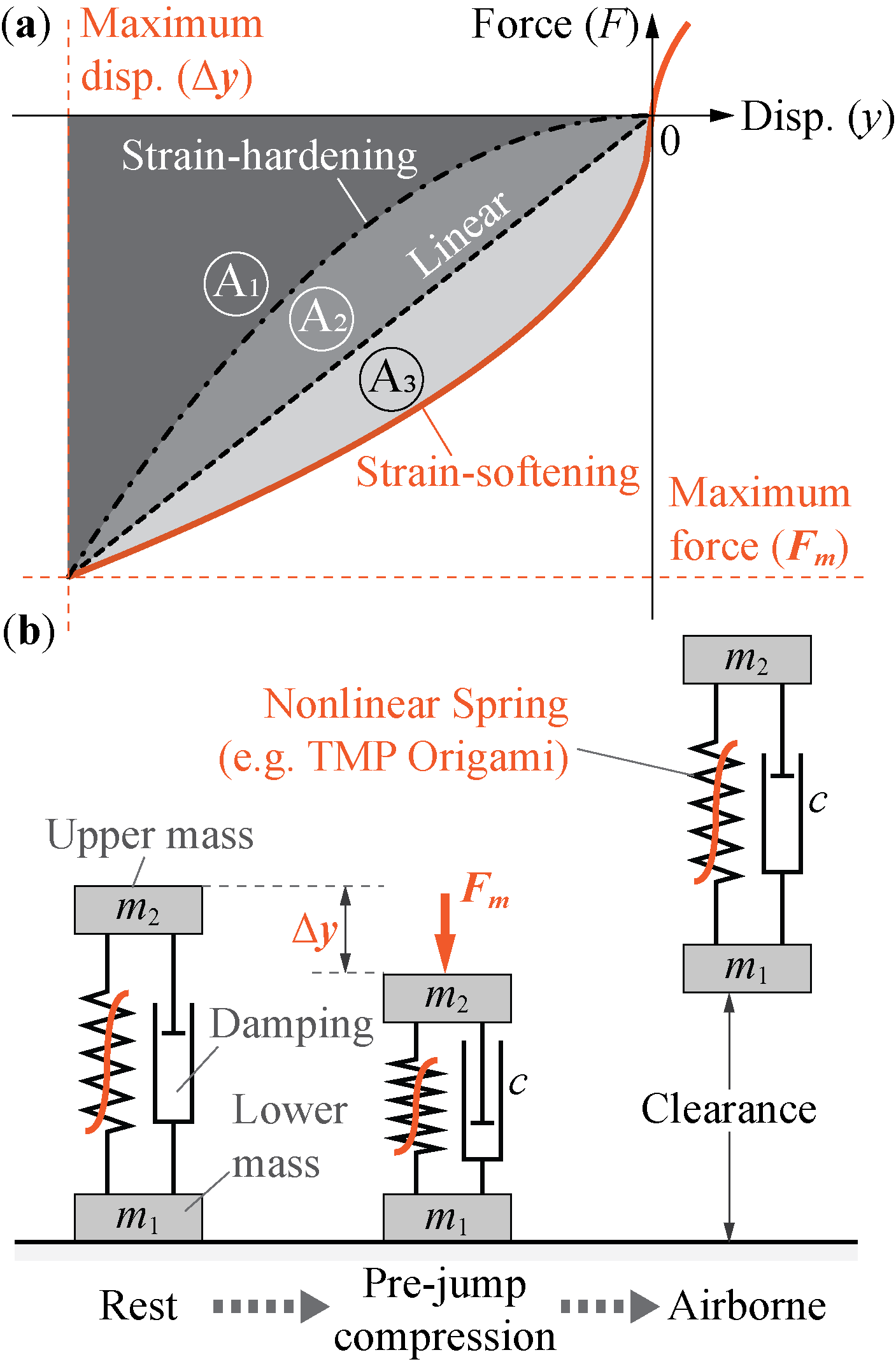}
    \caption{The overall background of this study.  a) Compression force-displacement curves of three generic spring elements within the same actuation and displacement limits. They are strain-hardening, linear, and strain-softening.  The shaded areas ($A_1$, $A_1+A_2$, and $A_1+A_2+A_3$) represent the stored energy in these three elastic elements. b) The jumping problem of interest in this study: As a jumping mechanism is compressed to the force and deformation limits, its nonlinear spring (e.g., TMP origami) stores elastic energy and then releases it to initiate the jump.  since the clearance and airborne time are directly related to the amount of released energy from the spring, a strain-softening mechanism offers the potential for improved jumping performance.}
    \label{fig:Generic_FD}
\end{figure}

In a generic, theoretical study, we showed that using a nonlinear elastic element with ``strain-softening'' characteristics can increase the stored energy pre-jump, and consequently offers higher jumps and ground clearance with a negligible penalty in efficiency \cite{sadeghi}.  Figure \ref{fig:Generic_FD} represents three different generic force-displacement curves with the same maximum compression force ($F_m$) and displacement ($\Delta y$).  They are strain-hardening, linear, and strain-softening, respectively.  The shaded areas, $A_1$, $A_1+A_2$, and $A_1+A_2+A_3$ represent the amount of stored energy in these three elastic elements. The strain-softening element clearly has the best performance regarding energy storage. 

While promising, the previous study was theoretical.  Therefore, this study aims to investigate the feasibility of using origami --- as a jumping robotic skeleton with embedded energy storage --- to generate the desired strain-softening behavior, and experimentally validate the corresponding improvement in jumping performance.   Origami has found many creative uses in robotics because it is elegant, simple to fabricate, and capable of achieving complex 3D shapes from an easy-to-transport flat configuration \cite{felton, lee2013deformable, shigemune2016origami, belke2017mori}.  Moreover, folding can also create unique and nonlinear elastic properties like tunable nonlinear stiffness \cite{fang,li} and multi-stability \cite{fang2, kamrava2019origami,sadeghi2, zhakypov2019designing, sadeghi2020dynamic}.  Tapping these folding-induced mechanics for robotic applications is still a nascent field full of promising potentials. 

\begin{figure}[h!]
    \hspace{-0.50in}
    \includegraphics[scale=1.0]{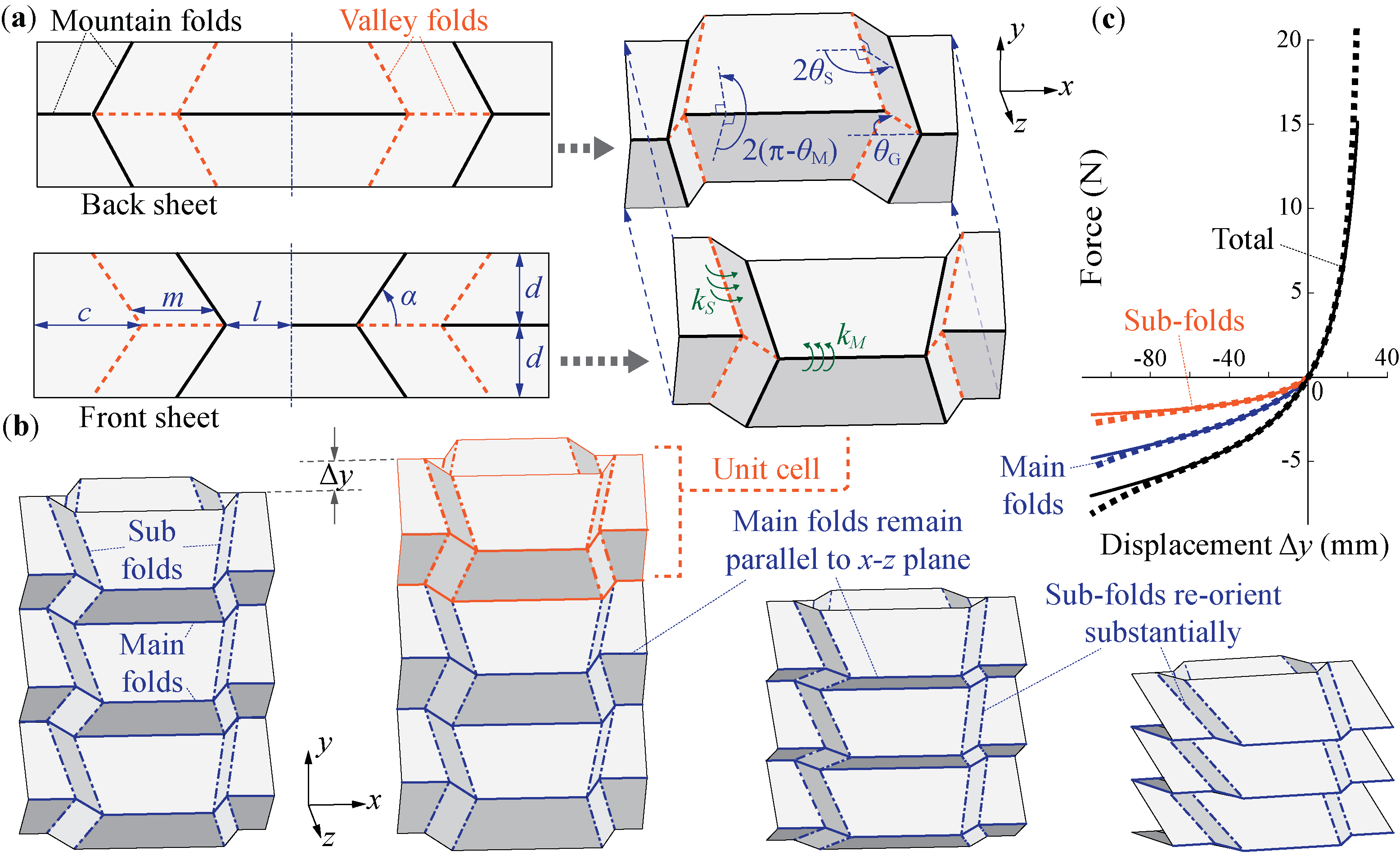}
    \caption{The design and mechanics modeling of Tachi-Miura Polyhedron (TMP) bellow. a) The crease pattern of TMP and the corresponding design and kinematics parameters.  b) TMP bellows at (from left to right) resting folding angle ($\theta_M=60\degree$), upper bound of the linear crease stiffness ($\theta_M=\theta_2=70\degree$), lower bound of linear crease stiffness ($\theta_M=\theta_1=38\degree$), and a highly folded configuration ($\theta_M=25\degree$). Note that the main folds refer to those parallel to the TMP base and sub-folds refer to those with an angle $\alpha$ to the main folds. c) The reaction force-displacement of a sample TMP design.  The desired strain-softening behavior is evident when $\Delta y<0$.}
    \label{fig:TMPgeometry}
\end{figure}

To this end, we use a modified Tachi-Miura Polyhedron (TMP) bellow origami (Figure \ref{fig:TMPgeometry}) as a nonlinear spring connecting two end masses that simulate payloads.  We used a rigid-folding based mechanics model to uncover the correlation between the nonlinear stiffness and TMP crease design, and this model is also used to simulate the jumping dynamics.  In this study, we use airtime (aka. the airborne time when the lower mass is above the ground) and clearance (aka. maximum height of the lower mass) as the performance metrics to assess the jumping performance and optimize the TMP design.  We also use a creasing technique known as plastically annealed lamina emergent origami (PALEO) \cite{klett} to minimize the hysteresis observed during the pre-jump compression and release to increase the jumping efficiency.  In addition to an TMP jumper with nonlinear strain-softening, we also fabricated a jumper designed with a close-to-linear force-displacement relationship. We experimentally compare the dynamic jumping performance of both designs.

In what follows, section 2 of this paper reviews the kinematics of TMP bellow, discusses its nonlinear stiffness modeling, and details the fabrication method to minimize hysteresis.  Section 3 presents the dynamic modeling of the jumping mechanism consisting of a TMP bellow and two end-point masses, and details the the results of the jumping experiments. Section 4 presents a parametric study on the effects of TMP design parameters on its nonlinear stiffness and jumping performance.  Finally, section 5 concludes this paper with a summary and discussion. 

\section{Using TMP Folding to Generate Strain-Softening} \label{Section:2}
\subsection*{Kinematics and Nonlinear Stiffness of Tachi-Miura Polyhedron Bellow} \label{Section:2-Subsection:1}

The Tachi-Miura Polyhedron (TMP) origami bellow, initially proposed in 2009 \cite{tachi2009one}, is a variation of the classical Miura-ori pattern. Each TMP bellow is a linear assembly of unit cells, and each cell consists of a front sheet and back sheet (Figure \ref{fig:TMPgeometry}(a)).  The TMP folding pattern includes horizontal ``main-folds'', as well as ``sub-folds'' that are creased at an angle to the main-folds.  During folding, the main-folds remain parallel to the TMP base (aka. The $x-z$ horizontal reference plane), but the sub-folds change their spatial orientation significantly (Figure \ref{fig:TMPgeometry}(b)).  TMP is rigid-foldable so its folding induce minimal deformation in the facets \cite{yasuda,liu}. This characteristic indicates that the structure's folding behavior can be described primarily by the deformation along the main-folds and sub-folds that behave like hinges with assigned torsional spring stiffness \cite{liu}. Using this rigid-foldable condition and the virtual work principle, we can calculate the TMP's force-displacement relationship as a function of dihedral crease angles \cite{yasuda}: 

\begin{equation}\label{eq:1}
F=\frac{-32}{d\cos\theta_M}\left[k_M\frac{N-1}{N}\left(\theta_M-\theta_{M_0}\right)+k_S\frac{\cos^3\frac{\theta_G}{2}\sin\theta_M}{\cos\alpha\sin\theta_S}\left(\theta_S-\theta_{S_0}\right)\right],
\end{equation}
where \textit{N} is the number of unit cells in the TMP. \textit{d} is the unit cell height.  $k_M$ and $k_S$ are the torsional spring coefficients of the main-folds and the sub-folds, respectively.  $\alpha$ is the angle of the sub-folds relative to the main-folds at the unfolded flat configuration. $\theta_M$ and $\theta_S$ are half of the dihedral angle of the main-folds and sub-folds, respectively. $\theta_{M_0}$  and $\theta_{S_0}$ are their corresponding resting angles. $\theta_G$ is the angle between the $x$-axis and the main fold (\ref{fig:TMPgeometry}(a)).  The kinematic relationships between these folding angles are: 

\begin{align}
\label{eq:2}
\theta_M &=\sin^{-1}\left(\sin\theta_{M_0}-\frac{\Delta y}{Nd}\right)\\
\label{eq:3}
\theta_G & =2\tan^{-1}\left(\tan\alpha\cos\theta_M\right)\\
\label{eq:4}
\theta_S &=\cos^{-1}\left(\frac{\sin 0.5\theta_G}{\sin\alpha}\right)
\end{align}
where $\Delta y$ is the deformation of the TMP from its resting height. Moreover, the crease torsional stiffness can be formulated as $k_M=\hat{k}_M\left(l+m+c\right)$ and $k_S=\hat{k}_S d\sin^{-1} \alpha$, where $\hat{k}_M$ and $\hat{k}_S$ are the torsional stiffness \emph{per unit length} of the main and sub-folds, respectively.  Finally, we define a folding ratio $R_F$ to determine the pre-jump TMP deformation through the final main-fold angle $\theta_{M_f}$,

\begin{equation}\label{eq:7}
    R_F=\frac{90\degree-\theta_{M_f}}{90\degree}
\end{equation}

However, the force-displacement relationship presented in Equation \eqref{eq:1} does not consider the deformation limits due to rigid folding. As the TMP is nearly compressed ($\theta_M=\theta_S \approx 0\degree$), its facets will come into contact with one another, increasing the reaction force significantly.  On the other hand, when the TMP is nearly stretched flat ($\theta_M=\theta_S \approx 90\degree$), its reaction force also increases significantly due to the in-plane facet stretching.  To more accurately calculate the force-displacement relationship, we impose an increase in main and sub-fold's torsional stiffness when their folding angles exceed pre-defined upper and lower bounds \cite{liu}.  Based on experimental data (as we detail later), we choose a lower bound of $\theta_1=38\degree$ and an upper bound of $\theta_2=70\degree$ to modify the $k_{M}$ and $k_{S}$ in Equation \eqref{eq:1}, as follows: 

\begin{equation}\label{modkM}
    k_M = 
    \begin{cases} 
    k_{M_0}\sec^2\left(\frac{\pi\left(\theta_M-\theta_1\right)}{3.5\theta_1}\right),        & \quad 0<\theta_M<\theta_1 \\
    k_{M_0},                                                                                & \quad \theta_1\leq\theta_M\leq\theta_2\\
    k_{M_0}\sec^2\left(\frac{\pi\left(\theta_M-\theta_2\right)}{2\pi-3.5\theta_2}\right),   & \quad \theta_2<\theta_M<\pi
    \end{cases}
\end{equation}

\begin{equation}\label{modkS}
    k_M = 
    \begin{cases} 
    k_{S_0}\sec^2\left(\frac{\pi\left(\theta_S-\theta_1\right)}{3.5\theta_1}\right),        & \quad 0<\theta_S<\theta_1 \\
    k_{S_0},                                                                                & \quad \theta_1\leq\theta_S\leq\theta_2 \\
    k_{S_0}\sec^2\left(\frac{\pi\left(\theta_S-\theta_2\right)}{2\pi-3.5\theta_2}\right),   & \quad \theta_2<\theta_S<\pi
    \end{cases}
\end{equation}

Figure \ref{fig:TMPgeometry}(c) shows the force-displacement curve of a baseline TMP design before and after the torsional stiffness modification ($c=d=l=m=30$mm, $N=8$, $\alpha=40\degree$, and $\hat{k}_M=\hat{k}_S=0.005\text{N/rad}$).  Overall, the TMP clearly shows the desired strain-softening nonlinearity \emph{in compression}.  It is worth noting that the TMP stiffness in tension has relatively minor effects on the jumping performance \cite{sadeghi}. To gain a better understanding, we separate the contribution from main and sub-folds in compression.  Although the main and sub-folds have the same torsional stiffness \emph{per unit lengnth}, they have different lengths (90mm for the main-fold and 46.6mm for the sub-folds). As a result, the main folds have a large contribution to the overall reaction force.  However, it is also evident from Figure \ref{fig:TMPgeometry}(c) that the response of the sub-folds exhibits a stronger nonlinearity than the main-folds.  At the beginning of compression, the sub-folds rest at an angle to the TMP base, significantly stiffening the overall structure.  As the TMP deforms in compression, the sub-folds gradually re-orient themselves and become more parallel to the horizontal main-folds, offering less resistance to folding (Figure \ref{fig:TMPgeometry}(b)).  On the other hand, the main-fold remain parallel to the horizontal reference plane throughout the folding process, so their response is relatively linear.  Therefore, to achieve a stronger strain-softening nonlinear stiffness, one should increase the sub-fold's torsional stiffness per unit length and soften the main folds. 

\subsection*{Prototyping for Minimal Energy Loss}
\begin{figure}[t]
    \hspace{-0.75in}
    \includegraphics[scale=1.0]{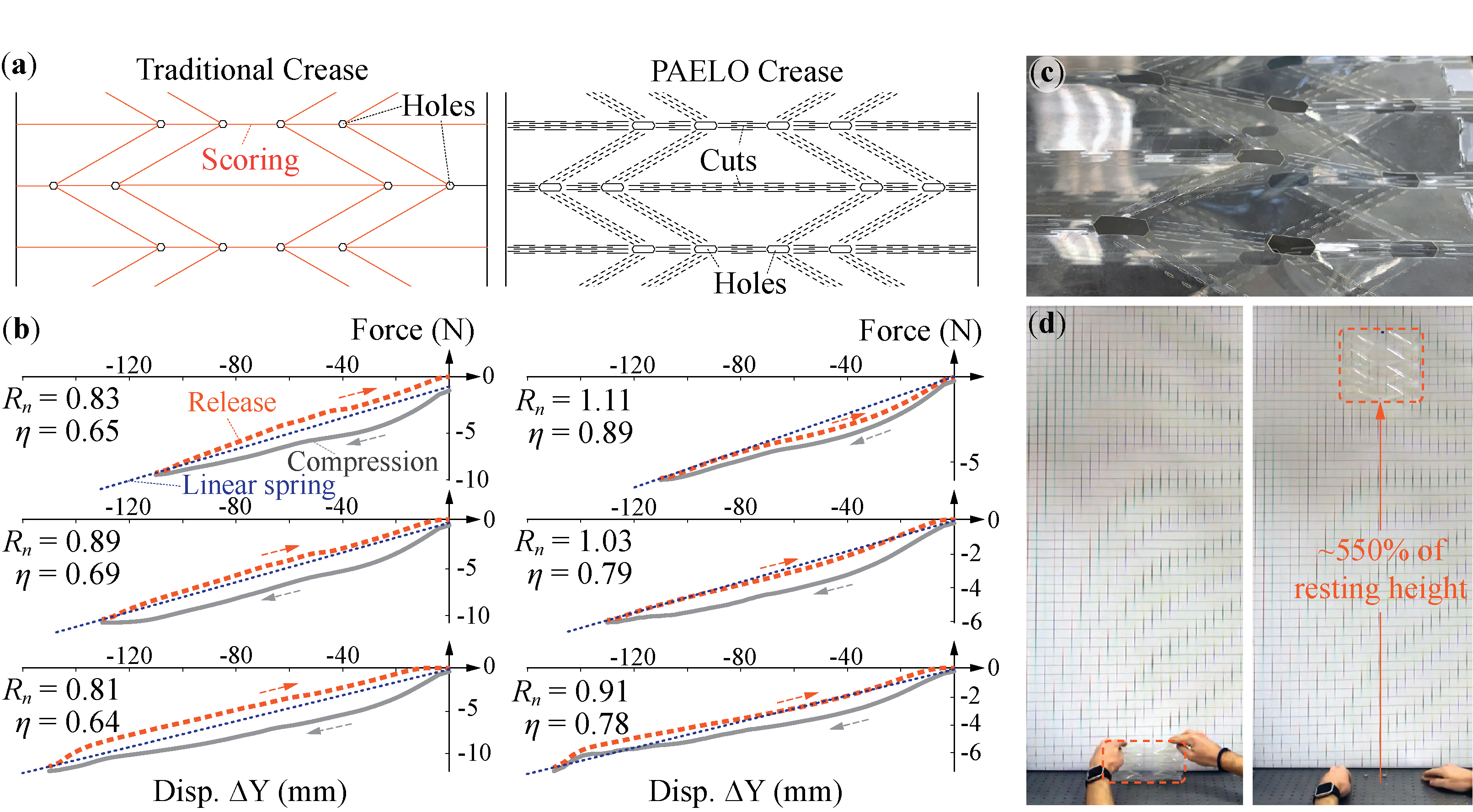}
    \caption{Fabrication with the PALEO technique to minimize the hysteresis. a) Left: A TMP crease pattern based on the traditional creasing method by scoring.  Right: The same TMP design with the PALEO creasing method.  Notice the polygonal holes at the vertices to eliminate stress concentration.  b) The averaged force-displacement curves of the two TMP prototypes at different maximum compression displacement. The corresponding nonlinearity ratio ($R_n$) and efficiency ratio ($\eta$) are summarized.  c) A closed-up look of the laser-cut PALEO crease showing the relatively smooth curvature near the fold lines. d) The result of a validation jump without any end masses (or payloads).}
    \label{fig:Fab}
\end{figure}

The analytical results in the previous subsection validate the feasibility of generating strain-softening compression nonlinear stiffness from TMP origami. However, it does not consider a critical issue: plastic deformation and energy loss due to creasing and folding.  In the classic origami, creases are created by concentrated material deformation along a thin line, which then holds its shape --- this methodology inherently creates plastic deformation and is not conducive to a springy-like response for efficient jumping.  For example, we fabricated a TMP bellow using 125 $\mu$m thick Polyethylene terephthalate (PET) sheets and classical creasing method (Figure \ref{fig:Fab}(a), see Table \ref{tab:TMPParameters} for the design parameters of this TMP).  We place the PET sheet on a plotter cutter (Graphtec FCX4000) and score the creases at a depth of 60 $\mu$m, or just under half the sheet thickness.  The creased plastic sheets are then folded and assembled into the bellow, which is then annealed at 170$\degree$C for 60 minutes to hold its shape.  We place the completed TMP bellow in a universal tester machine (ADMET eXpert 5061 with 25 lbs load cell), compress it to a maximum displacement of 110mm, 130mm, and 150mm (or 75\%, 79\% and 85\% of folding ratio), and then release it to simulate the pre-jump compression stage illustrated in Figure \ref{fig:Generic_FD}(b).  The end openings of TMP bellow change their shape during folding, so we placed Teflon sheets on the contact surface between TMP and tester machines to minimize sliding friction.  Figure \ref{fig:Fab}(b) summarizes the averaged compression force-displacement curves from five loading cycles.  It is evident that, during compression, the TMP exhibits the desired strain-softening behavior.  However, during the release stage, the strain-softening no longer exists, and the TMP bellow indeed releases less energy than an equivalent, perfectly linear spring that has the same reaction force at the maximum compression.  Such a significant hysteresis behavior is a result of the concentrated plastic deformation along the crease lines, and it negates the potential benefits from strain-softening since the jumping performance is only related to the released energy.

To quantitatively characterize the performance of TMP regarding energy release, we define two different metrics.  The first metric --- referred to as the ``efficiency ratio'' $\eta$ --- is the ratio of the released energy from TMP during relaxation over the stored energy during compression.  

\begin{equation}
\label{eq:19}
\eta=\frac {E_\text{release}}{E_\text{compress}}.
\end{equation}

A higher efficiency ratio close to one is desirable for minimal energy loss due to hysteresis. The second metric --- ``nonlinearity ratio'' $R_n$ --- is the ratio of the \emph{released} energy from TMP over the potential energy of an equivalent, perfectly linear spring (aka. a spring that has the same reaction force as the TMP at the maximum compression).   To benefit from the nonlinear stiffness for better jumping performance, $R_n$ should be larger than one \cite{sadeghi}.   However, TMP with the classic creasing fails to meet this target in that the efficiency ratio is low around 0.65, and the corresponding nonliearity ratio upon release is consistently below 1. 

To address the issues of high hysteresis and low nonlinearity ratio during release, we use a creasing technique known as plastically annealed lamina emergent origami (PALEO).  PALEO works by folding the creased polymer sheet into its desired state --- often by plastically deforming the material --- and then annealing it well above the glass transition temperature to relax the internal stresses (170$\degree$C for 60 minutes, then a rapid cooling of 15$\degree$C/min to the glass transition temperature, and finally a slow cooling of 0.5$\degree$C/min to room temperature) \cite{klett, sargent}. This procedure leaves the creases in a stress-free state and allows the material to remain in the elastic range during folding, thus minimizing hysteresis.  In addition to annealing, PALEO creases are designed with a series of cuts parallel to the crease (Epilog Fusion M2 laser cutter, Figure \ref{fig:Fab}(c)) so that one can fine-tune the crease torsional stiffness per unit length ($\hat{k}_M$ and $\hat{k}_S$) by adjusting the length and spacing between these small cuts.  In this way, we can apply a stiff PALEO crease to the sub-folds and soft crease to the main-folds to amplify the nonlinear strain-softening.   

Figure \ref{fig:Fab}(b) displays the same TMP design with the PALEO crease and the corresponding force-displacement curves. (Section 1 of the Appendix details the PALEO design and its corresponding torsional stiffness measurement.)  It is evident that the PALEO design significantly reduces the hysteresis (aka. higher efficiency ratio of $\eta$).  The cuts incorporated into the PALEO creases can increase the bend radius along the main-folds and sub-folds, thus reducing the amount of energy loss due to plastic deformation (Figure \ref{fig:Fab}(c)). More importantly, the more efficient PALEO creases can increase the nonlearity ratio and make the TMP origami outperform the equivalent linear spring during the releasing stage.  However, these benefits become less evident as the maximum displacement increases.  Therefore, a compression displacement of 110 mm was selected for subsequent dynamic jumping tests in an effort to maximize the benefits of nonlinearity.  In a validation jump, the TMP bellow with PALEO crease, without end masses to simulate the payload, reached more than 500\% of its resting height (Figure \ref{fig:Fab}(d) and Supplemental Video 1).

\begin{table}[t]
    
    \caption{Design parameters, crease stiffness, and the two energy metrics of the nonlinear and linear TMP prototypes. The unit of $c,$ $d,$ $l,$ and $m$ is millimeter, and the unit of $\hat{k}_M$ and $\hat{k}_S$ is N/rad.}
    \label{tab:TMPParameters}
    %\hspace{-0.25in}
    \scalebox{0.9}{
    \begin{tabular}{*{11}{c}}
        \hline
         & \multicolumn{6}{c}{Design parameters} & \multicolumn{2}{c}{Crease stiffness} & \multicolumn{2}{c}{Energy metrics}\\
         \cline{2-11}
         & $N$ & $\alpha$ & $c$ & $d$ & $l$ & $m$ & $\hat{k}_{M_0}$ & $\hat{k}_{S_0}$ & $R_n$ & $\eta$\\
         Nonlinear & 8 & $30\degree$ & 21.1 & 34.2 & 39.9 & 23.2 & 0.0186 & 0.0946 & 1.12 & 0.89\\
         Linear & 8 & $50\degree$ & 34.0 & 21.6 & 31.7 & 39.1 & 0.0946 & 0.0186 & 1.02  & 0.85\\
         \hline
    \end{tabular}
    }
\end{table}

\begin{figure}[t]
    \centering
    \includegraphics[scale=1]{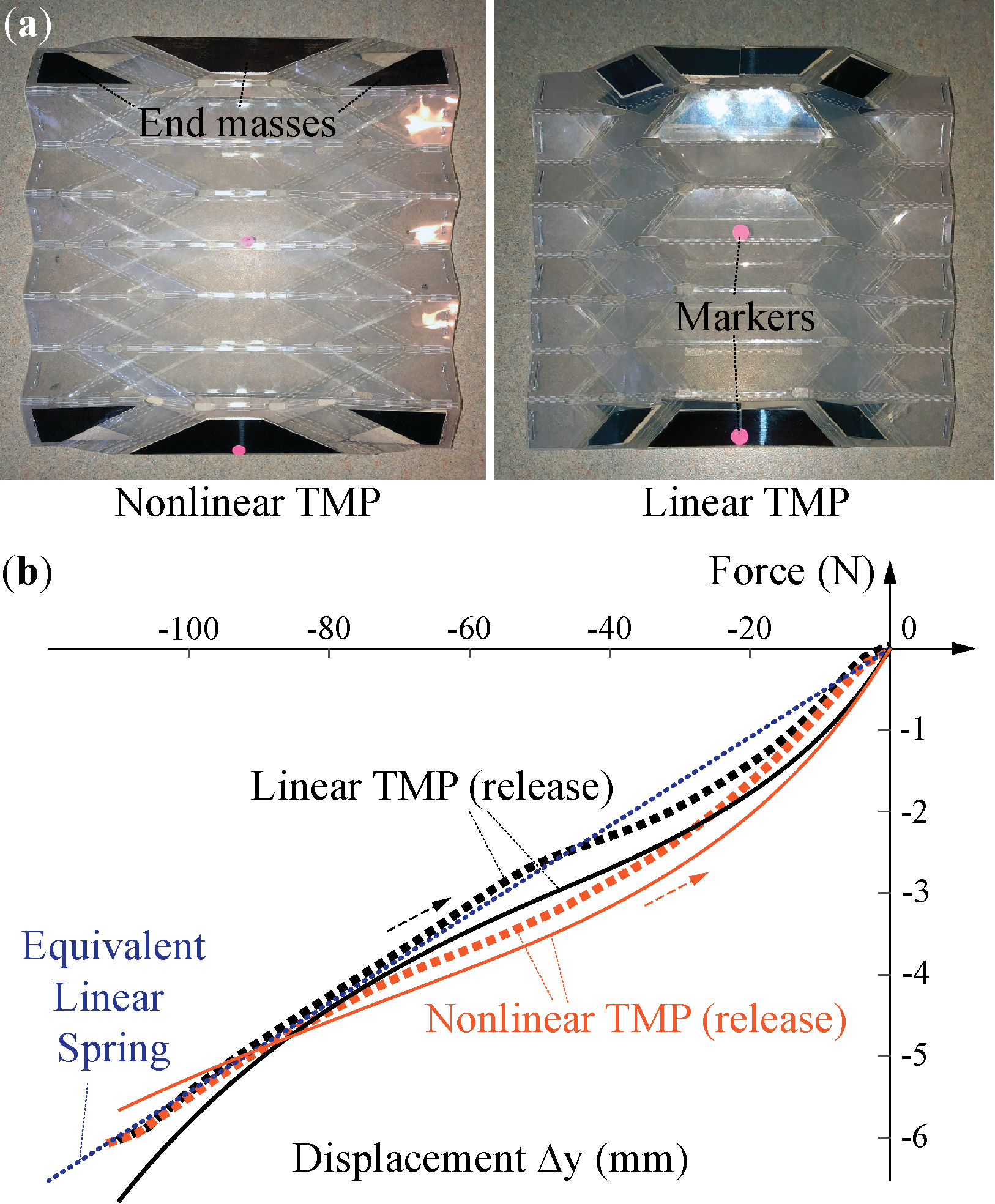}
    \caption{Experimental force-displacement curves of nonlinear and linear TMP in compression. The dashed lines are measured force-displacement curves \emph{during the releasing stage} (as indicated by the dashed arrow) and thin solid lines are the corresponding theoretical predictions.  Notice that the measured curves have the same reaction force at the maximum compression (110mm), which are used to construct the equivalent linear spring.}
    \label{fig:LinearTMP}
\end{figure}

To validate that strain-softening nonlinearity can improve the jumping performance, we fabricate an additional TMP prototype with a close-to-linear force-displacement relationship during the pre-jump release.  This linear TMP has the same reaction force as the nonlinear one at 110mm compression displacement (Figure \ref{fig:LinearTMP}).  It also features the PALEO crease design but has stiff main-folds and soft sub-folds to reduce the nonlinearity.  The linear TMP geometric design parameters, summarized in Table \ref{tab:TMPParameters}, are obtained and modified from an optimization (technical details of this optimization and the modification we made are detailed in the Section 2 of Appendix).

Figure \ref{fig:LinearTMP} compares the averaged compression force-displacement relationships of both nonlinear and linear TMP samples based on five consecutive loading cycles.  The nonlinearity ratio of the linear TMP is $R=1.02$, indicating that it is a good representation of an equivalent, ideally linear spring. In comparison, the nonlinear TMP sample achieved a nonlinearity ratio of $R=1.13$. The percent error between the theoretical and experimental results represents the quality of agreement between the force-displacement model described in Section \ref{Section:2-Subsection:1} and the experimental results. The model has excellent agreement with the nonlinear TMP at 2.77\% error regarding the amount of released energy, while the linear TMP shows acceptable agreement at 11.34\% error. The effect of the error in the approximation of energy storage will be discussed later. 

\section{Dynamic Jumping of TMP Mechanism}\label{Section:3}

\subsection*{Numerical Simulation}
\begin{figure}[h!]
    \hspace{-0.75in}
    \includegraphics[scale=1.0]{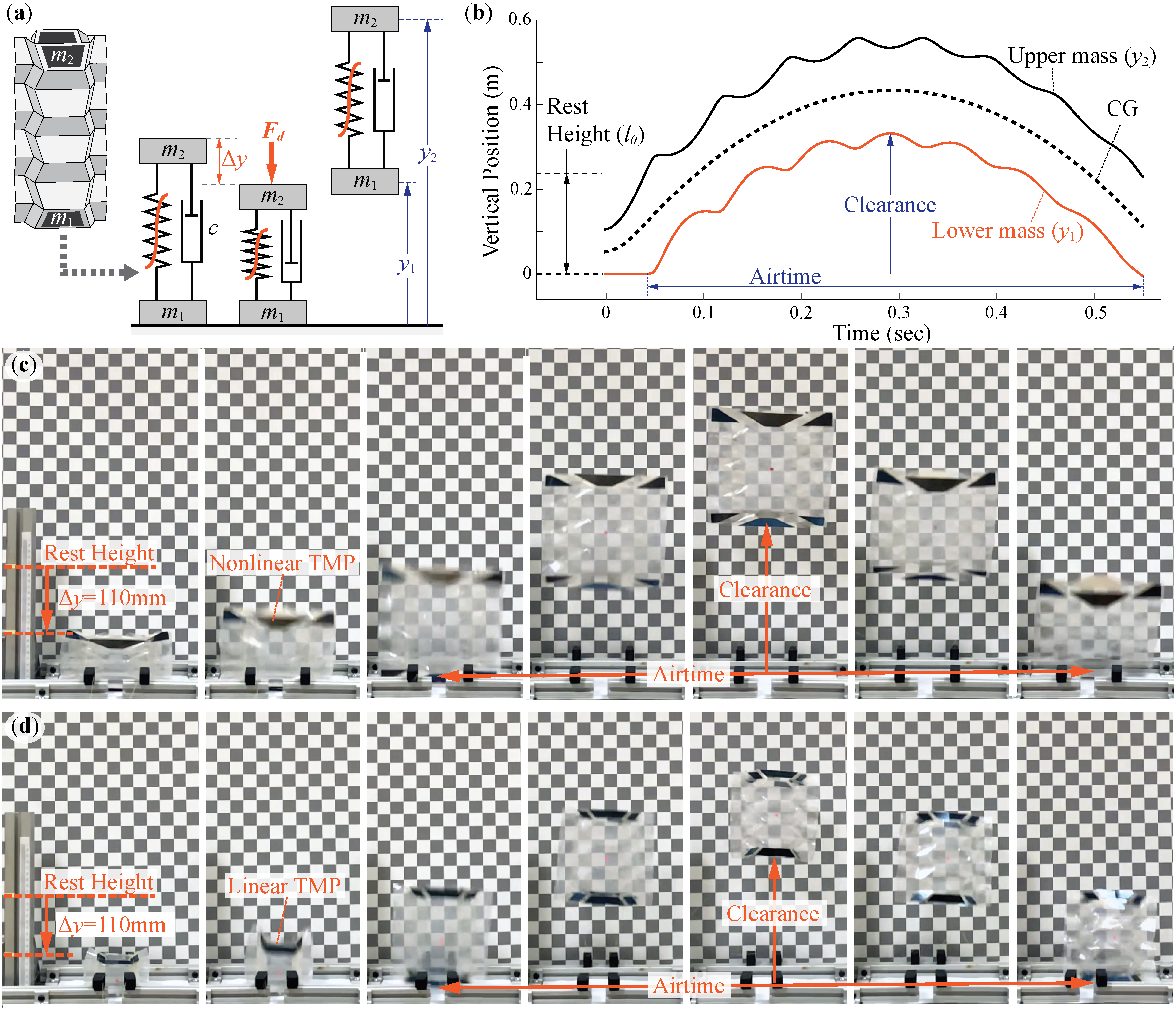}
    \caption{Dynamic jumping test. a) Schematic of the dynamic setup. b) A simulated time-response of the upper and lower mass of the structure as well as the center of gravity (CG). Dynamic jumping performance metrics of the mechanism (aka. clearance and airtime) are shown. c, d) Video footage of the jumping tests of the nonlinear and linear TMP, respectively.}
    \label{fig:Time_response}
\end{figure}

Figure \ref{fig:Time_response}(a) illustrates the dynamic setup of the TMP jumping.  We secure thin steel plates to the facets at the two ends of TMP bellow to simulate the payloads, so that the overall system dynamics is approximately equivalent to that of two end masses connected by a nonlinear spring and damping. 

The jumper is actuated by an external force applied on the upper mass ($m_2$), which compresses the TMP and stores energy, then this force is released instantaneously to initiate jumping.  The jumping includes two consecutive phases: pre-jump and post-jump.  During the pre-jump phase, the upper mass ($m_2$) accelerates upward due to the TMP's reaction force, but the lower mass ($m_1$) remains on the ground.   Due to the complexity of implementing structural damping in dynamic simulation, we use equivalent linear viscous damping instead, so the equation of motion during the pre-jump phase is,

\begin{equation}
    \label{eq:10}
    m_2\ddot{y}_2=F(y_2-l_0)-m_2g-C\dot{y}_2
\end{equation}
where $F$ is the reaction force from the TMP bellow as a function of relative displacement between the two end masses (based on the nonlinear folding mechanics model in the previous section).  $C$ is the equivalent damping coefficient, which is related to the experimentally observed hysteresis in that \cite{blandon}

\begin{equation}
    \label{eq:13}
    C=2(m_1+m_2)\zeta_{eq}\omega_{n,eq}
\end{equation}
where the equivalent damping ratio $\zeta_{eq}=0.25(1-\eta)/\pi$, the equivalent natural frequency 

\begin{equation}
    \label{eq:12}
    \omega_{n,eq}=\sqrt{\frac{k_{eq}(m_1+m_2)}{m_1m_2}}
\end{equation}
and $k_{eq}$ is the spring coefficient of the equivalent, perfectly linear spring.

The post-jump phase starts when the lower mass ($m_1$) leaves the ground and ends when it returns to the ground.  In order for the lower mass to leave the ground, the tensile reaction force from the TMP bellow must exceed the weight of the lower mass: $F>m_1g$. The equations of motion in this phase are:

\begin{align}
     m_1\ddot{y}_1  &=-F(y_2-y_1-l_0)-m_1g+C\left(\dot{y}_2-\dot{y}_1\right)  \\
     m_2\ddot{y}_2  &=F(y_2-y_1-l_0)-m_2g-C\left(\dot{y}_2-\dot{y}_1\right)
\end{align}

The pre-jump phase results are used as the initial conditions for the governing equations during the post-jump phase.  Figure \ref{fig:Time_response}(b) shows the simulated jumping response of both TMP jumper and the equivalent, perfectly linear jumper. 

\subsection*{Jumping Performance Assessment} \label{Section:2-Subsection:3}

To accurately represent end-point masses on the TMP jumpers, we waterjet cut steel shim stocks to size and adhered them to the top and bottom facets using double-sided tape (Figure \ref{fig:LinearTMP}).  The geometry and number of these shim stocks are chosen carefully to ensure the same end masses on the two TMP jumpers for direct performance comparison. In the completed prototypes, the lower mass $m_1$ is 17.7g for the nonlinear TMP and 17.9g for the linear TMP, and the upper mass $m_2$ is  17.6g and 17.9g, respectively.

To conduct the jumping test, we place the jumper in front of a 1-inch square grid paper, providing a height reference for clearance data.  We use two thread segments secured to the jumping platform just behind the TMP sample to achieve the initial compression.  We drape these threads over the TMP sample, pull them down manually, compress the TMP to 110mm displacement, and then simultaneously cut the two threads to release the TMP jumper.  Dynamic data was collected by video recording each jump at 240 frames per second and analyzing the footage, frame-by-frame, to extract take-off time, landing time, and maximum clearance of the lower mass.  Figure \ref{fig:Time_response}(c, d) shows the still images of two jumping trials.  The airtime is the difference between the time stamp on the ``take-off frame'' (Frame 3 in figure \ref{fig:Time_response}) and the ``landing frame'' (Frame 7).  The clearance was measured using the reference grid, as shown in Frame 5 of Figure \ref{fig:Time_response}(c,d). 

\begin{table}[]
    \centering\
    \caption{Jumping performance of the nonlinear and linear TMPs.}
    \label{tab:JumpPerformance}
    \scalebox{0.9}{
    \begin{tabular}{c c c c c}
    \hline
         & \multicolumn{2}{c}{Nonlinear} & \multicolumn{2}{c}{Linear}\\
         \cline{2-5}
         & Experiment & Analytical & Experiment & Analytical \\
         \cline{2-5}
         Air-time [sec] & 0.461 $\pm$ 0.014 & 0.495 & 0.423 $\pm$ 0.011 & 0.484\\
         Clearance [mm] & 259.7 $\pm$ 14.3  & 314.8 & 230.6 $\pm$ 14.0 & 293.7\\
         \hline
    \end{tabular}
    }
\end{table}

Table \ref{tab:JumpPerformance} summarizes the averaged results of ten jumping trials of both TMP jumpers.  Both TMP show consistent performance, and more importantly, the nonlinear TMP shows a 9\% improvement in airtime and 13\% improvement in clearance over the linear TMP.  The experimentally measured jumping performance metrics are consistent with the corresponding theoretical predictions (also summarized in Table \ref{tab:JumpPerformance}).  The discrepancy in airtime prediction is acceptable at under 10\% for the nonlinear TMP and under 15\% for the linear TMP.  However, the discrepancy in ground clearance is more significant, especially for the linear TMP.  Besides the error in force-displacement prediction discussed in Figure \ref{fig:LinearTMP}, another potential cause for such discrepancy is the lateral oscillation of the end mass in addition to the vertical oscillation.  This lateral oscillation is particularly evident in the linear TMP (supplemental video 2).  Regardless, the experiment results validate our hypothesis that a ``strain-softening'' nonlinearity can lead to more stored energy pre-jump and improve the jumping performance compared to a perfectly linear mechanism. 

\section{Correlating the TMP Design to Jumping Performance} \label{Section:2-Subsection:2}

To understand how the TMP folding geometry influences the nonlinear strain-softening stiffness and jumping performance, we perform a parametric study by varying the design variables discussed in the previous sections, one at a time, between two pre-defined limits (Table \ref{tab:ParameterSweepTable}).  These upper and lower limits are chosen based on folding kinematics, fabrication, and many other factors.  For example, we set the lower limit of the unit cell number ($N$) at 6 to reduce unfavorable boundary effects and set the upper limits at 10 to avoid localized buckling during compression.  The choice of the lower limit on the sub-fold angle $\alpha$ can also help avoid buckling during compression (a small $\alpha$ angle will result in a thin TMP in the z-axis), and the upper limit can ensure proper foldability. The upper limit of the length parameters $d$, $l$, $m$, and $c$ are determined based on the fabrication capability available to the authors, and the lower limit ensures sufficient space for PALEO creasing and proper folding.  It is worth emphasizing that the sub-folds in all these TMP designs use the stiff PALEO crease design (Table \ref{tab:PALEOtable}), and the main-fold uses the soft crease design to maximize the nonlinearity.

\begin{figure}[t]
    \hspace{-0.6in}
    \includegraphics[scale=1.0]{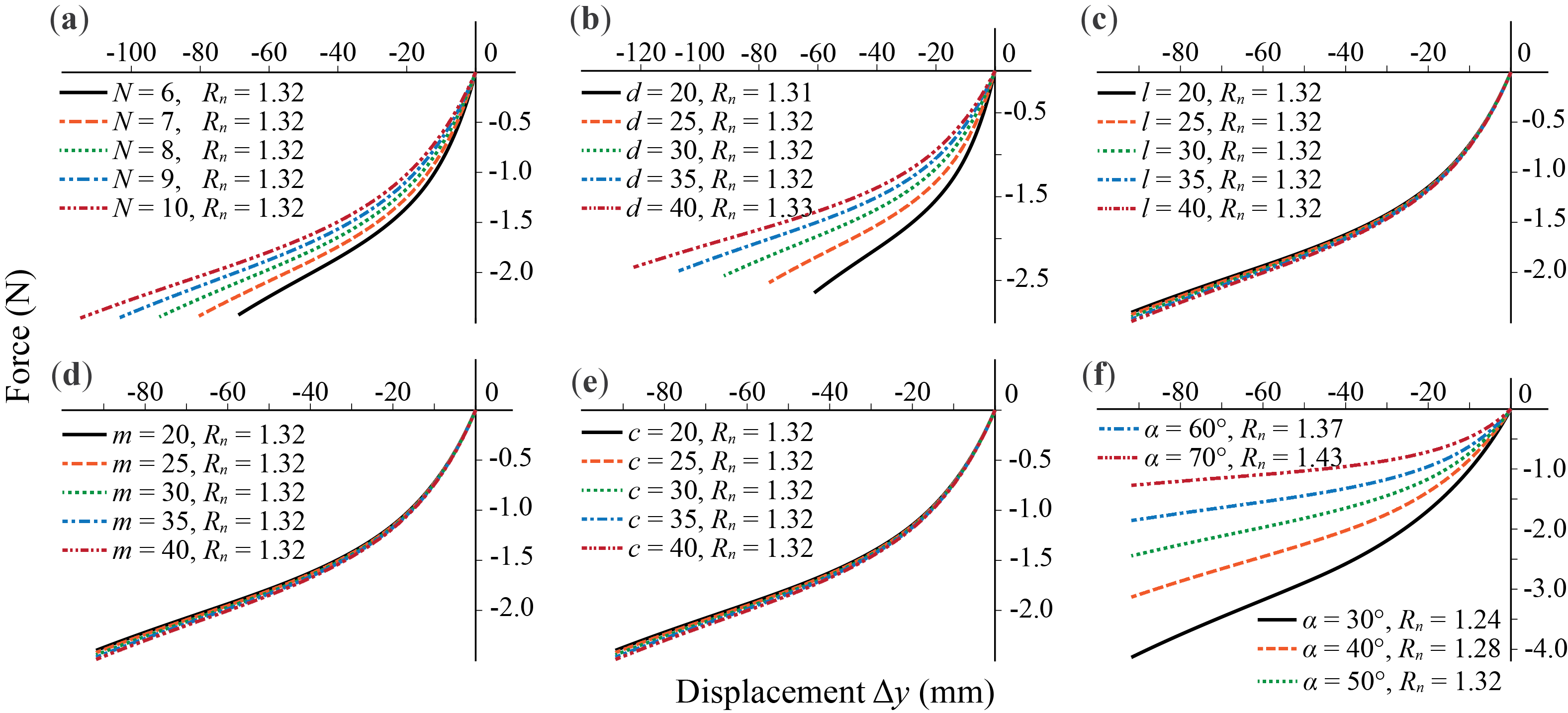}
    \caption{Parametric study of the TMP design parameters' effect on the force-displacement relationships. The unit of the crease length variables ($d$, $l$, $m$, $c$) is millimeter.}
    \label{fig:Parameter_sweep}
\end{figure}

Figure \ref{fig:Parameter_sweep} summarizes the compression force-displacement curves of different TMP designs and their corresponding nonlinear ratio.  It is important to note that for each design, the maximum displacement corresponds to a folding ratio of $R_F=0.75$.  Since the folding ratio bases on the changes in the crease angles (Equation \ref{eq:7}), using it to calculate the maximum displacement can avoid excessive plastic deformation and hysteresis loss.  In other words, if the final displacement were constant for all TMP designs, those with smaller resting height (e.g., TMP designs with small $N$ and $d$) would undergo excessive folding that creates significant energy loss during the compression stage. 

\begin{table}
\centering
\caption{TMP design variables used for the parametric study}
    \scalebox{0.9}{
\begin{tabular}{c c}
     \hline
     Base Values & Limits for parameter study \\
     \hline
     $N=8$ & $6<N<10$ \\
     $d=30\text{mm}$ & $20\text{mm}<d<40\text{mm}$ \\
     $1=30\text{mm}$ & $20\text{mm}<1<40\text{mm}$ \\
     $m=30\text{mm}$ & $20\text{mm}<m<40\text{mm}$ \\
     $c=30\text{mm}$ & $20\text{mm}<c<40\text{mm}$ \\
     $\alpha=50\degree$ & $30\degree<\alpha<70\degree$ \\
     \hline
\end{tabular}\label{tab:ParameterSweepTable}
    }
\end{table}

Based on the results in Figure \ref{fig:Parameter_sweep}, it is immediately evident that some parameters have more significant influences on the TMP's nonlinearity than others.  The main-folds' lengths (\textit{l}, \textit{m}, and \textit{c}) have little impact on the force-displacement curve because of the soft PALEO crease design.  The unit cell number ($ N $) affects the TMP's reaction force magnitude but will not affect the nonlinearity ratio.  Increasing the unit cell height $d$ (which essentially increases the stiff sub-fold's length) can increase the reaction force's magnitude; however, the corresponding increase in the nonlinearity ratio is small.  The sub-fold angle $\alpha$ has a strong influence on both the reaction force magnitude and the nonlinearity ratio.  A higher $\alpha$ angle would give a higher nonlinearity ratio $R_n$ but a smaller reaction force magnitude at the same compression displacement.  This is because a higher $\alpha$ angle makes the sub-folds more perpendicular to the TMP base before folding, thus amplifying their spatial re-orientation during folding (Figure \ref{fig:TMPgeometry}) and increasing the nonlinearity ratio. On the other hand, a higher $\alpha$ angle decreases the stiff sub-fold's length, thus reducing the TMP's overall stiffness.  

Based on these parametric study results, we further examine the correlation between the sensitive TMP design parameters (aka. unit cell number $N$, unit cell height $d$, and sub-fold angle $\alpha$) and the dynamic jumping performance by numerical simulations. 

\begin{figure}[t]
    \hspace{-0.6in}
    \includegraphics[scale=1.0]{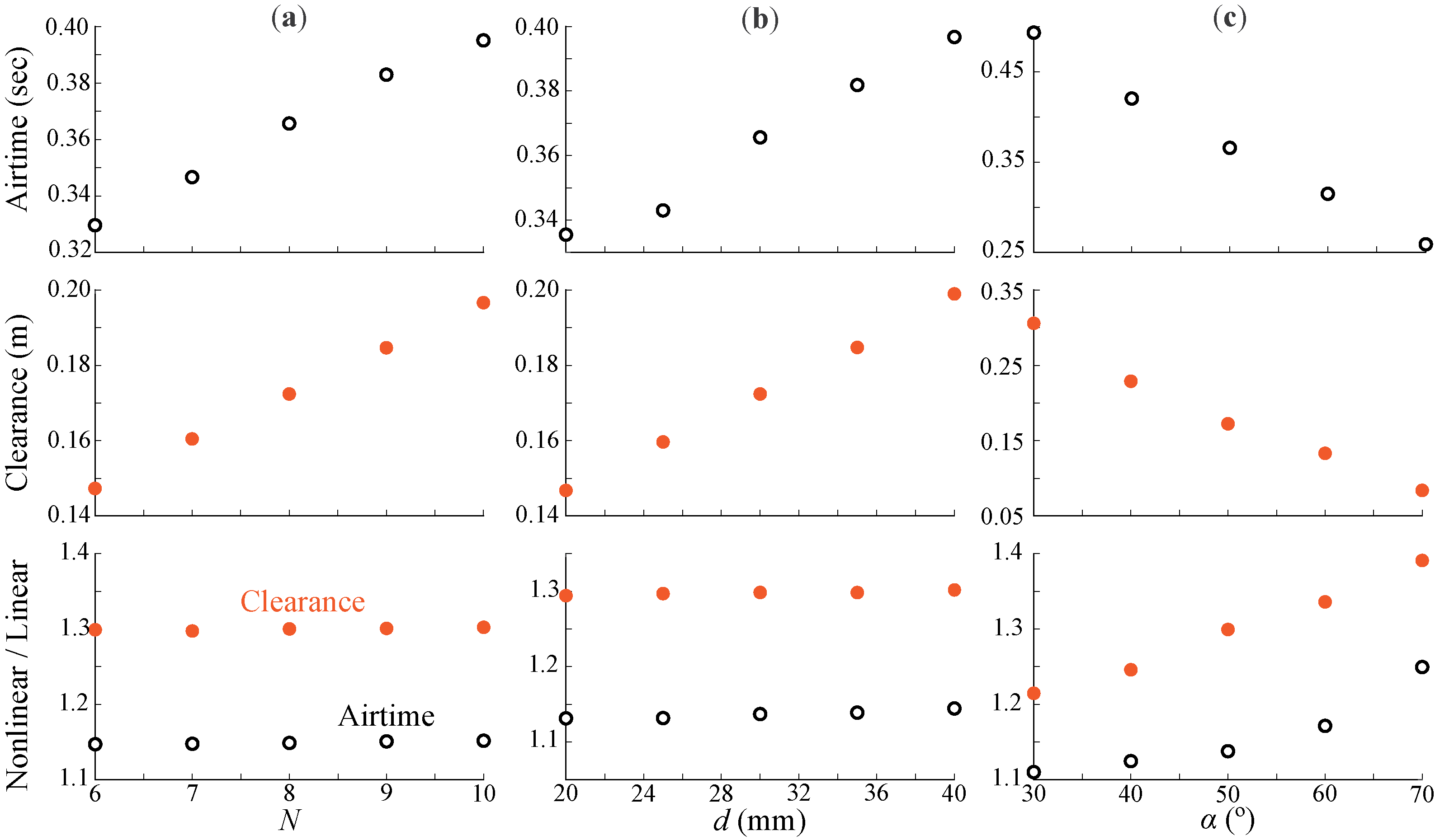}
    \caption{Parametric study on the correlation between TMP's jumping performance and its a) unit cell number ($N$), b) unit cell height ($d$), and c) sub-fold angle ($\alpha$).  The first row shows the airtime performance, and the second rows shows the clearance. The third row shows the airtime and clearance ratio between the nonlinear TMP bellows and their corresponding, equivalent linear spring.}
    \label{fig:Dynamic_Parameter_sweep}
\end{figure}

We observe that a TMP with a larger number of unit cells undergoes larger deformation before reaching the folding limit of $R_F=0.75$, which results in a more considerable amount of stored energy upon release.  As a result, increasing $N$ gives a higher airtime and clearance (Figure \ref{fig:Dynamic_Parameter_sweep}(a)).   A TMP with a higher unit cell length has a higher reaction force under the same compression deformation, thus giving higher airtime and clearance as well (Figure \ref{fig:Dynamic_Parameter_sweep}(b)).  However, these two design parameters have no or minimal effects on the nonlinearity ratio.  As a result, even though TMPs will outperform their equivalent linear springs ($R_n>1$ for all considered designs), such out-performance does not change with $N$ or $d$.

The angle of the sub-folds $\alpha$, on the other hand, affects both reaction force magnitude and nonlinearity ratio.  When $\alpha$ angle increases, the TMP outperforms its equivalent linear spring more because of a higher nonlinearity ratio; however, the absolute values of airtime and clearance decreases due to a smaller reaction force and a smaller amount of stored energy (Figure \ref{fig:Dynamic_Parameter_sweep}(c)).  This tradeoff indicates that design optimization would be necessary when the TMP is used in robotic applications subjected to different compression forces and displacements constraints. 

\section{Summary and Conclusion}
\label{Section:5}

This study demonstrates the effectiveness of using TMP origami to design and fabricate a jumping mechanism that outperforms a linearly elastic element.  We begin with a theoretical study of the nonlinear stiffness of TMP bellows by assuming rigid facets and hinge-like fold lines with nonlinear torsional spring coefficients.  The theoretical prediction shows that TMP bellows with appropriate designs exhibit a ``strain-softening'' nonlinear stiffness, thus storing more elastic energy for jumping than an equivalent linear spring with the same reaction force at maximum compression.  This nonlinearity originates from the sub-folds' spatial re-orientation during folding; thus, stiff sub-folds and soft main-folds are desirable.

However, we encounter a significant challenge when fabricating a TMP jumper that can harness this nonlinearity for better dynamic performance:  the hysteresis due to the concentrated plastic deformation along the fold lines.  This hysteresis significantly reduces the released energy from TMP right before jumping.  To address this issue, we adopt a PALEO creasing design technique, which places carefully spaced cuts (perforations) along the fold lines and then applies annealing after folding to release the residual stress.  Moreover, PALEO creasing allows us to fine-tune the crease torsional stiffness for the main-folds and sub-folds.  We then fabricate a nonlinear and close-to-linear TMP bellow with the same reaction force at maximum compression and conduct dynamic jumping tests.  The results show that the nonlinear TMP outperforms the linear one in terms of air time and clearance. This out-performance is quantitatively consistent with the numerical simulations that treat the TMP as a nonlinear spring connecting two end masses. 

Finally, parametric studies on both force-displacement curves and dynamic jumping performance indicate that the number of unit cells, unit cell height, and sub-fold angles will affect TMP's jumping performance. Increasing unit cell number and height can offer higher airtime and clearance, but it has minimal effects on the overall nonlinearity.  On the other hand, changing the sub-fold angle will create a tradeoff between the airtime/clearance and nonlinearity ratio.  These parametric studies offer the necessary design insights for optimizing the TMP bellow for future robotic applications with different constraints. 

Origami has its inherent advantages for robotic applications (e.g., easy to fabricate, compact in the folded configuration, and scalable in size). This study elucidates a new dimension of using the mechanics of folding for robotic applications, thus expanding the tools available to robotics engineers.

\section{Acknowledgement}
The authors thank Dr. Phanindra Tallapradaga at the Clemson University for thoughtful discussions at the beginning of this project. The authors acknowledge the support from the National Science Foundation (CMMI 1751449 CAREER and 1933124).  S. Li also acknowledge the additional support from Clemson University (via startup funds and CECAS Dean's Faculty Fellow Award).

\bibliographystyle{elsarticle-num-names}
\bibliography{sample.bib}

\newpage

\begin{figure}[h!]
    \hspace{-0.25in}
    \includegraphics[scale=0.99]{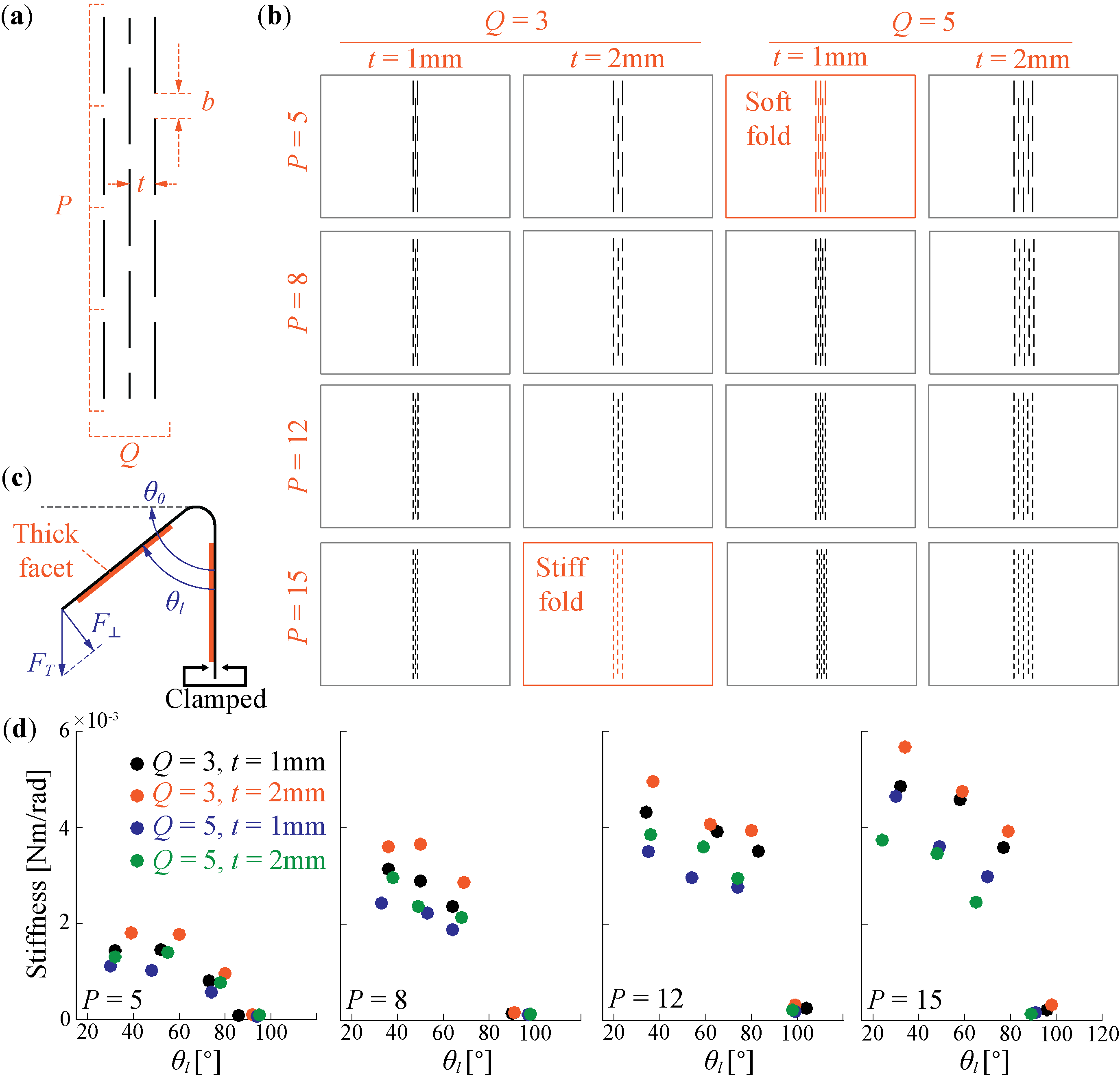}
    \caption{Design and testing of PALEO crease. a) The design parameters of PALEO crease. b) The sixteen different PALEO designs tested for this study, the ``stiff'' and ``soft'' crease designs used for the TMP prototyping are highlighted. c) The schematic showing the test setup for torsional stifffness.  d) A summary of the test results.}
    \label{fig:PALEO}
\end{figure}

\section{Appendix} \label{Section:Appendix}

\subsection{PALEO Crease Stiffness} \label{Section:Appendix_Subsection:3}
To investigate how the cut geometry affects the PALEO crease stiffness, we fabricated sixteen samples with different combinations of cut length ($P$), the spacing between cuts ($t$ and $b$), and the number of cut lines ($Q$) (Figure \ref{fig:PALEO}(a, b)).   After cutting, we fold these samples completely until their facets come to contact, and then placed them into a customized frame at a 90$\degree$ dihedral angle for annealing.  After annealing, we reinforced the facets on either side of the crease with a 1.5 mm thick layer of polypropylene to prevent facet deformation during testing.  The stiffness of each crease sample was measured by applying a load to the stiff facet's endpoint and observing the crease's angular deflection (Figure \ref{fig:PALEO}(c)).  We take four measurements with increasing loads to understand how the crease stiffness changes through the folding angle from 90$\degree$ to roughly 30$\degree$.  Torsional stiffness is calculated using the equation below, 

\begin{equation}
    \label{eq:20}
    k_{\theta}=\frac{F_{\perp}d_{\perp}}{\theta_0-\theta_l},
\end{equation}
where $F_\perp$ is the perpendicular component of the applied load, $d_\perp$ is the perpendicular distance from the crease line to the applied load position, and $\theta_0$ and $\theta_l$ are the resting and the angle after loading, respectively.  The results of the stiffness study are summarized in Figure \ref{fig:PALEO}(d).

The results show that the torsional stiffness is relatively small near the resting angle ($\theta_0=90\degree$) but quickly increases as the folding angle decreases. This trend supports the assumption that the crease stiffness should be increased as the folding angle becomes small, as shown in Equations \eqref{modkM} and \eqref{modkS}.  Table \ref{tab:PALEOtable} summarizes the cutting pattern selected for the ``stiff'' and ``soft'' creases in the nonlinear and linear TMP prototype, along with the corresponding torsional spring stiffness \emph{per unit length}.  The nonlinear TMP prototype uses the stiff PALEO design for its sub folds and soft design for the main folds.  On the other hand, the linear TMP sample uses the stiff design for its main folds and soft one for the sub folds.

\begin{table}
    \centering
    \caption{Selection of PALEO crease configurations. Here, $k_{\theta}$ is the torsional  stiffness of the tested sample, and $\hat{k}_{\theta}$ is the corresponding torsional stiffness per unit length.}
    \label{tab:PALEOtable}
    \scalebox{0.9}{
    \begin{tabular}{c c c c c c c}
    \hline
     & $P$ & $Q$ & $t$ & $b$ & $k_{\theta}$ & $\hat{k}_{\theta}$ \\
     \hline
     Soft fold  & 5 & 5 & 1mm & 5mm & 0.0011 Nm/rad & 0.0186 N/rad\\
     Stiff fold & 15 & 3 & 2mm & 2mm & 0.0057 Nm/rad & 0.0946 N/rad\\
     \hline
    \end{tabular}
    }
\end{table}

\subsection{Design of Linear TMP}
\label{Section:Appendix-Subsection:2}
To experimentally validate the benefits of using a nonlinear elastic element to enhance the jumping performance, we designed a different TMP prototype showing a close-to-linear force-displacement response.  This linear TMP's design parameters were determined using the Simplex and Powell single objective optimization algorithms in the ModeFRONTIER$^\text{TM}$. We use two different algorithms as a way to cross-validate the results of the optimization.

To design the linear TMP, we first define an ideally linear force-displacement curve based on the final compression force and displacement of the nonlinear TMP as the objective.  The optimization is then used to minimize the total error between the linear TMP force-displacement curve and this ideally linear response (Equation \eqref{eq:18}). 

\begin{equation}
    \label{eq:18}
    \text{Minimize:}\; e=\sum \mid{F_{TMP}-F_{ideal}}\mid
\end{equation}

\begin{table}[t]
    \centering
    \caption{Linear TMP parameter sets resulting from single objective optimization, the units of $c$, $d$, $l$, and $m$ are millimeters.}
    \label{tab:LinearParameterSets}
    \scalebox{0.9}{
    \begin{tabular}{*{8}{c}}
         \hline
         & \multicolumn{6}{c}{Design Variables} & \\
         \cline{2-7}
         & $N$ & $\alpha$ & $c$ & $d$ & $l$ & $m$ & Error [N]\\
         Powell & 8 & $70\degree$ & 34.0 & 21.6 & 31.7 & 39.1 & 205.1 \\
         Simplex & 8 & $70\degree$ & 39.6 & 21.1 & 20.0 & 40.0 & 206.5 \\
         \hline
    \end{tabular}
    }
\end{table}

More specifically, we calculate the error between these two curves at roughly 2000 sample points within the displacement range, and the sum of these errors is the objective function.  The upper and lower bounds placed on the linear TMP design variables are consistent with those shown in Table \ref{tab:ParameterSweepTable}.  The crease torsional stiffness of its main fold is based on the stiff PALEO crease design in that $(\hat{k}_M=0.0946 \frac{\text{N}}{\text{rad}})$ (Table \ref{tab:PALEOtable}), and the sub fold uses the soft PALEO design $(\hat{k}_S=0.0186 \frac{\text{N}}{\text{rad}})$.  More importantly, we directly constrain the final compression force and displacement of the linear TMP to be the same as those of nonlinear TMP prototype, allowing for a more direct comparison.  

Table \ref{tab:LinearParameterSets} shows the optimized linear TMP design parameters resulting from both Powell and Simplex algorithms. Both methods yield similar magnitudes of error, but the Powell algorithm results show slightly better agreement with the ideally linear spring.  However, the optimized linear TMP parameters resulted in difficulties during experimental testing.  The large sub-fold angle $\alpha = 70\degree$ gives undesirable compression behavior in that the TMP structure buckled rapidly cell-by-cell under increasing compression, generating an inconsistent force response curve.  Through trial and error, we lower the sub-fold angle until smooth compression cycles could be achieved consistently.  And the modified sub-fold angle is $\alpha = 50\degree$.  The modified linear TMP design is summarized in Table \ref{tab:TMPParameters}.

\end{document}